# Multiple Categories Of Visual Smoke Detection Database

Yafei Gong
Faculty of Information Technology,
Beijing University of Technology

Xuanchao Ma
Faculty of Information Technology,
Beijing University of Technology

*Abstract*—Smoke detection has become a significant task in associated industries due to the close relationship between the petrochemical industry's smoke emission and its safety production and environmental damage. There are several production situations in the real industrial production environment, including complete combustion of exhaust gas, inadequate combustion of exhaust gas, direct emission of exhaust gas, etc. We discovered that the datasets used in previous research work can only determine whether smoke is present or not, not its type. That is, the dataset's category does not map to the real-world production situations, which are not conducive to the precise regulation of the production system. As a result, we created a multi-categories smoke detection database that includes a total of 70196 images. We further employed multiple models to conduct the experiment on the proposed database, the results show that the performance of the current algorithms needs to be improved and demonstrate the effectiveness of the proposed database.

## I. INTRODUCTION

Petrochemical companies are the primary producers of energy in China, and smoke detection at petrochemical companies has long drawn considerable interest. It is necessary that exhaust gas emitted by the venting flare must be entirely burnt to guarantee the long-term normal operation of the system and efficiently eliminate environmental pollution. The main existing solution to reduce exhaust gas pollution emissions is to inject combustion-supporting steam to promote the complete combustion of exhaust gas [1]. With the proper amount of combustion-supporting steam, exhaust gases can be completely burned to reduce pollution. If the amount of combustion-supporting steam is inadequate, the smoke produced by incomplete combustion will pollute the air. However, when the amount of combustion-supporting steam is excessive, it not only results in a waste of resources but also obstructs the exhaust gas combustion, resulting in the direct release of the exhaust gas into the atmosphere. Therefore, the key to the highly effective combustion of exhaust gas is the suitable regulation of combustion-supporting steam. Current approaches rely based on sensors and human adjustment, but due to their unique features, it is impossible to ensure the effective and efficient combustion of exhaust gas using these two methods [2]-[4]. With the advancement of artificial intelligence, image-based technologies may offer novel solutions to the task of smoke detection.

The state of combustion of exhaust gas can be directly reflected in the image. During the combustion of exhaust gas, the color of smoke will vary according to the degree of combustion and will also represent the industrial environment's operating conditions. For instance, in petrochemical plants and refineries, the presence of the smokeless state indicates that the system is operating normally. When black smoke is detected, the system is in abnormal operating, indicating that there is insufficient combustion of the exhaust gases. The higher the carbon content, the darker the hue, and the larger the environmental harm, the blacker the smoke will be. The white smoke indicates excessive combustion-supporting steam, and the exhaust gas is released straight into the atmosphere, which will seriously pollute the environment. Consequently, when black smoke or smokeless is discovered, it may cause significant safety dangers as well as environmental issues.

In order to reduce pollution, particulate matter detection [5]–[10] and image-based smoke detection methods have been the subject of extensive research [11]–[15]. As far as we are aware, the available methods with their data sets can only determine whether smoke is present or absent, they cannot determine the type of smoke [16]-[20]. Furthermore, actual industrial control systems can only make basic decisions based on limited smoke detection results, and the new algorithms developed using these datasets are unable to provide additional information to assist the control systems in producing more precise regulation. As a result, creating a database with multiple categories has a significant realistic impact on resolving the aforementioned engineering issues. Therefore, we proposed and created a multi-classification database for smoke detection in this paper with a total of 70196 image patches.

The remaining structure of this paper is organized as follows. The process for creating the dataset is described in depth in Section 2. The effectiveness of the database and the performance of the algorithm are examined in Section 3. The main conclusion is offered in Section 4 at the end.

## II. DATABASE

During the combustion of exhaust gases, three types of smoke are often produced: white smoke, black smoke, and smokeless, which is also the most common form of smoke. Similar findings have been seen in other industrial domains, such as thermal power plants. As a result, we create a new three-categories smoke detection database (TCSDD) for smoke detection based on [15] to continue the research on the aforementioned engineering difficulties.

The dataset includes a training set, a validation set, and a testing set in accordance with standard dataset design practices. In the process of designing the database, it is found that the color of smoke in an image does not exactly match pure black or white. In order to identify the categories of an image according to the color of smoke, the subjective scoring technique is utilized. An image's label should be derived from the outcomes of multiple labeling to minimize the learning error brought on by inaccurate labels. This study uses 20 data from 20 different sources to determine an image's label. The selection of the final label from a range of results can be equivalent to the hard voting process in ensemble learning. To be specific, let $X=[x_1,x_2,...,x_n]$, where n is the number of categories and $x_i$ is the number of votes an image has received to be classified in a certain category (n equals three in this paper). The sum of $x_i$ should also equal M, where M is the total number of votes cast for an image, which is 20. The following is the image's final category:

$$x = argmax(X) \qquad (1)$$

*argmax* denotes that the final category of the image is obtained from the subscript of the value that causes *X* to obtain the maximum value.

Another issue that was discovered while building the database is the stark variation in the number of images in each category. The performance of the network depends on the distribution of samples among the categories [20]. Specifically, there are 8363 smokeless images, 1423 black smoke images, and 778 white smoke images in the validation set that was initially partitioned, compared to 8511 smokeless images, 1423 black smoke images, and 778 white smoke images in the initially built training set. As a result, data augmentation is needed for images of white smoke and black smoke. Considering the color change enhancement has significant interference with the smoke image, that should be considered when developing the specific algorithm. Therefore, the rotation operation is mostly employed to augment data. The rotation angle that is applied to the image can be calculated by 360 / N * I, where N is the multiple of the data increase and I = [1,2,..., N]. For the black smoke images, the rotation operation with n equals 6 is utilized, and for the white smoke images, the rotation operation with n equals 12 is used, which closely balances the number of images in the three categories. By utilizing rotation, the total number of images in the three categories is almost equal.

The finished dataset is as follows: The training set consists of 26538 images, comprising 9336 image patches of white smoke, 8511 image patches of smokeless, and 8538 image patches of black smoke. A total of 26483 image patches, 8363 smokeless image patches, 8928 black smoke image patches, and 9192 white smoke image patches make up the validation set. There are a total of 17328 image patches in the testing set, including 9888 image patches of smokeless patches, 4644 image patches of black smoke, and 2796 image patches of white smoke. The images in the dataset are displayed in Figure 1, it is important to note that only the images from the training set and the validation set are rotated.

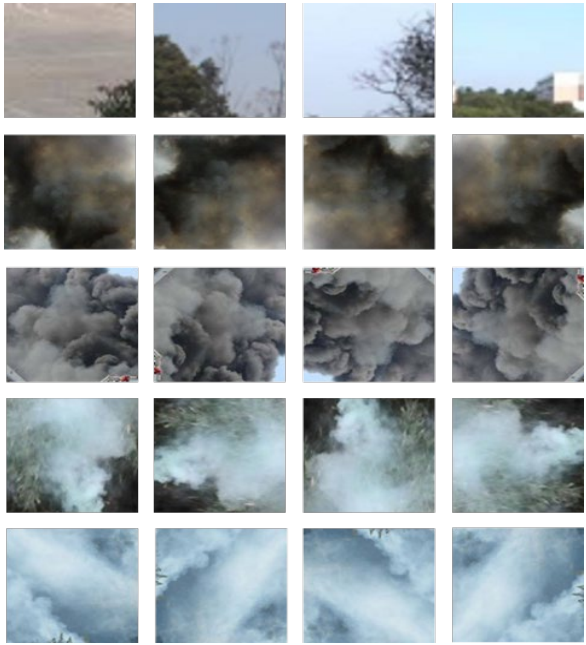

Figure 1 Examples of data, the patches from left to right are the original image and the image enhanced by rotation.

## III. EXPERIMENT

In this section, we will further examine the performance of existing state-of-the-art general classification networks and networks designed specifically for smoke detection, as well as assess the usefulness of this database.

### A. Experimental Protocol

**Evaluation Criteria.** For quantifying the performance of each model, The loss, precision, recall, and accuracy of the validation set and testing set, as well as the recall and precision of each category, were recorded. From these indicators, we were able to analyze the model's behavior and assess the database's efficacy. The following are the definitions of these indicators:

$$accuracy = \frac{TP+FP}{TP+TN+FP+FN} \quad (2)$$

$$precision = \frac{TP}{TP+FP} \quad (3)$$

$$recall = \frac{TP}{TP+FN} \quad (4)$$

$$f1socre = 2\frac{precision*recall}{precision+recall} \quad (5)$$

TP, TN, FP, and FN represent True positive, True negative, False Positive, and False negative respectively. In the binary classification task, recall and precision are typically calculated for the positive and negative samples. We determine the recall and precision for each category, then average the results to get the overall recall and precision. A good model is expected to achieve greater values on these indicators as far as possible.

**Operating Environment.** Operating Environment. All models used in this paper have the same configuration to allow for a fair comparison of model performance. Our experimental framework is PyTorch. An Ubuntu computer running the Inter(R) Gold 6248R CPU at 3.00GHz and an NVIDIA GeForce RTX 3090 graphics card power the experimental environment. All models have the same configuration for their hyperparameters. Table 1 displays the experiment's hyperparameter setting.

TABLE I. HYPERPARAMETER SETTINGS USED IN THE EXPERIMENT

|  | **Hyperparameter settings** |
|---|---|
| Epoch | 200 |
| Batch Size | 128 |
| Optimizer | SGD base_lr:0.01 Momentum:0.9 Weight_decay:1E-5 |
| Loss | CrossEntropyLoss |
| Lr Scheduler | Learning rate scales linearly from base_lr to 1E-5 |

**Competing Models.** The comparative experiments are conducted on our database with a total of seven start-of-the-art models, those models are Alex-Net [21], VGG-Net [22], Res-Net [23], Google-Net [24], Mobile-Net [25], Shuffle-Net [26], DCNN [15]. The first four networks have seen significant advancements in general image classification datasets in recent years and have been widely employed. MobileNet and ShuffleNet are created for real-world industrial uses. They are simpler to apply in the industry, having great precision and quick calculating speeds. DCNN is developed by Gu et al. specifically for smoke detection.

## B. Performance Comparison

On the TCSDD database, we record and compare the performance of the aforementioned seven start-of-the-art models using four indicators: accuracy, recall, precision, and f1-score. Loss is also employed as a measure of the robustness and generalization of networks. In this study, the loss of a model is determined on the testing set by averaging the total loss that results from adding the losses of all the samples in the testing set. The performance results of the network are shown in Table 2.

TABLE II. PERFORMANCE COMPARISON OF SEVEN STATE-OF-THE-ART

| Networks | AlexNet | ResNet | Vgg | GoogleNet | ShuffleNet | MobileNet | DCNN |
|---|---|---|---|---|---|---|---|
| Acc | 0.928 | 0.926 | 0.941 | **0.954** | 0.922 | 0.942 | 0.935 |
| R | 0.928 | 0.922 | 0.934 | **0.949** | 0.928 | 0.941 | 0.926 |
| P | 0.907 | 0.908 | 0.929 | **0.943** | 0.904 | 0.926 | 0.926 |
| F1 | 0.918 | 0.915 | 0.932 | **0.946** | 0.916 | 0.933 | 0.926 |
| loss | 0.328 | 0.380 | 0.895 | **0.191** | 0.343 | 0.276 | 0.358 |

Accuracy, precision, recall, and the f1-score are each denoted by the letters Acc, R, P, and F1 in Table 2. It can be shown from Table 2 that GoogleNet performs best, with an accuracy of 0.954, a recall of 0.949, and a precision of 0.943. Additionally, we discover that all models' precision is lower than recall, which indicates that the model's precision needs to be increased. In real-world applications, we not only need to be able to accurately identify particular scenarios, but we also don't want the model to become perplexed by various scenes. Therefore, it is necessary to work on increasing precision while keeping recall. For the indicators of each category, the experimental results demonstrate that the smokeless category has a higher recall and precision than the white smoke and black smoke categories, which is consistent with prior research's findings and suggests that the model does a good job of differentiating between smoke- and non-smoke-filled environments. In other words, it can effectively detect whether smoke is present in the image in the binary classification test. when employing TCSDD, the main errors occurred in the black smoke and white smoke categories. There has been a lot of work put into building the database to prevent importing inaccurate label information as much as possible. This demonstrates that there are additional causes for the uncertainty, like backdrop confusion in the image or color confusion brought on by changes in illumination. Features that are unaffected by environmental noise and brightness variations should be further mined by the models.

According to the above content, A basic ranking of the models may be determined based on the metrics, and the rankings are as follows: GoogleNet > MobileNet > VGG ≈ DCNN > AlexNet ≈ ResNet18 ≈ ShuffleNet. The best performance is achieved by Googlenet, this may be related to its multi-scale information extraction mechanism. VGG lags behind Mobilenet because its loss on the test set is substantially higher than Mobilenet's, suggesting that its generalization performance might be a little bit worse. Mobilenet also shows excellent capabilities, which also increases the potential for practical industrial applications. In addition, DCNN also offers great detection capabilities.

## IV. CONCLUSION

Smoke detection is essential to reducing air pollution and ensuring safe production. In this paper, we create a three-categories smoke database dedicated to petrochemical smoke detection. First, the image labels are obtained by the method of subjective scoring, then data augmentation is performed by rotation to solve the problem of category imbalance. The final generated dataset contains 70,196 image patches. we have run a performance competition on the proposed database with some state-of-the-art models, and the results show that performance still needs to be improved. In our future work, we will strive to leverage self-supervision and other techniques to mine smoke features to improve the performance of fully-supervised tasks for the challenge of distinguishing different forms of smoke.